\documentclass[conference]{IEEEtran}
\IEEEoverridecommandlockouts
\usepackage{cite}
\usepackage{amsmath,amssymb,amsfonts}
\usepackage{algorithmic}
\usepackage{graphicx}
\usepackage{textcomp}
\usepackage{xcolor}
\usepackage[colorlinks,citecolor=blue,urlcolor=blue,bookmarks=false,hypertexnames=true]{hyperref}
\usepackage{algorithm}
\usepackage[switch]{lineno}
\usepackage{paralist}
\usepackage{enumitem}

\usepackage{multirow}
\usepackage{makecell} %

\usepackage{booktabs} %
\usepackage{multirow} %
\usepackage{pgfplots} %
\usepackage{pgfplotstable} %
\usepackage{threeparttable}  %
\pgfplotsset{compat=1.18}
\usepackage{microtype}

\def\BibTeX{{\rm B\kern-.05em{\sc i\kern-.025em b}\kern-.08em
    T\kern-.1667em\lower.7ex\hbox{E}\kern-.125emX}}
\begin{document}

\title{Hierarchical Knowledge Structuring for Effective Federated Learning in Heterogeneous Environments
}

\author{
\IEEEauthorblockN{
Wai Fong Tam\IEEEauthorrefmark{1},
Qilei Li\IEEEauthorrefmark{1},
Ahmed M. Abdelmoniem\IEEEauthorrefmark{1}
}
\IEEEauthorblockA{\IEEEauthorrefmark{1}Queen Mary University of London, London, United Kingdom\\
\texttt{\{w.tam, q.li, ahmed.sayed\}@qmul.ac.uk}
\thanks{This work was supported by the UKRI and EPSRC under grant EP/X035085/1.}
}}

\maketitle

\begin{abstract}
Federated learning enables collaborative model training across distributed entities while maintaining individual data privacy. A key challenge in federated learning is balancing the personalization of models for local clients with generalization for the global model. Recent efforts leverage logit-based knowledge aggregation and distillation to overcome these issues. However, due to the non-IID nature of data across diverse clients and the imbalance in the client's data distribution, directly aggregating the logits often produces biased knowledge that fails to apply to individual clients and obstructs the convergence of local training. To solve this issue, we propose a Hierarchical Knowledge Structuring (HKS) framework that formulates sample logits into a multi-granularity codebook to represent logits from personalized per-sample insights to globalized per-class knowledge. The unsupervised bottom-up clustering method is leveraged to enable the global server to provide multi-granularity responses to local clients. These responses allow local training to integrate supervised learning objectives with global generalization constraints, which results in more robust representations and improved knowledge sharing in subsequent training rounds. The proposed framework's effectiveness is validated across various benchmarks and model architectures.
\end{abstract}


\section{Introduction}

Federated learning (FL) \cite{mcmahan2023communicationefficientlearningdeepnetworks} is a distributed machine learning approach under privacy protection. In FL, clients join forces to learn a global model without sending their data. In real-world applications, data distributions across clients are often non-independent and identically distributed (Non-IID). For instance, lighting conditions and weather can cause variations in the data collected by distributed video surveillance cameras \cite{kornblith2019similarityneuralnetworkrepresentations, 8100055, 9706717, MOU2021108038}. Conventional FL methods like FedAvg\cite{mcmahan2023communicationefficientlearningdeepnetworks} often falter, resulting in lower global models' performance under the non-IID data distributions \cite{zhao2018federated, zhu2021federatedlearningnoniiddata}, where data is not dispersed similarly among clients.  

The primary goals of alleviating this issue are improving model generalization to serve more clients or enhancing model personalization to fit local data distributions better. However, as local data distributions often differ from the global distribution in non-IID FL, these two optimal goals are usually at odds.

Researchers proposed Personalized Federated Learning (PFL) as an alternative to traditional Federated Learning (FL) to tackle the problems of statistical heterogeneity in local datasets (i.e., non-IID data), enabling clients to train models better suited to their data \cite{article}. Unlike conventional FL, which focuses on training a shared global model by minimizing the average loss across clients, PFL aims to create personalized models tailored to the unique data distribution of each client. PFL seeks a balance during training, i.e., personalized models must fit their local data distributions while benefiting from shared context knowledge obtained through collaborative training. However, a key challenge with PFL is enabling personalized models to learn unique patterns of the client's data and the shared knowledge between all clients to achieve a good combination of personalization and generalization. 

While recent PFL methods outperform traditional FL approaches, they often depend on a single global model to guide local training, which limits their ability to achieve effective personalization. The current PFL often derive personalized models from a global model trained by FedAvg \cite{9210355, sim2019investigationondevicepersonalizationendtoend}. However, if client data distributions are vastly different (i.e., non-IID), the global model may fail to represent all distributions effectively. Relying on a single global model can degrade the overall performance of personalized models. Therefore, balancing personalization and generalization is crucial, as local and out-of-distribution classes often appear during inference.

To address this, we introduce Hierarchical Knowledge Structuring (HKS), a framework that balances personalization and generalization in FL. In HKS, clients train models locally and share knowledge with the server, which organizes logits into a bottom-up hierarchical structure. Clients then utilize the aggregated logits at a granularity level suited to their tasks.

Our main contributions can be summarized as follows:
\begin{itemize}
    \item \textbf{Hierarchical Knowledge Distillation}: We propose a method that enhances both \textbf{global performance} and \textbf{personalization} by reinforcing knowledge exchange through hierarchical clustering, addressing the trade-off between generalization and personalization in FL.
    
    \item \textbf{Stronger Data Privacy-Preservation}: Unlike existing methods, HKS does not require clients to share data labels, effectively reducing privacy risks while enabling \textbf{personalized learning} via sharing logits with the server.
    
    \item \textbf{Effective Performance in Heterogeneous Settings}: Experimental results demonstrate that \textbf{HKS} outperforms state-of-the-art methods in personalized tasks, while maintaining \textbf{competitive global performance} in diverse non-IID scenarios.
\end{itemize}

\section{Related Work}

\subsection{Federated Learning}
Numerous FL strategies, such as FedAvg \cite{mcmahan2023communicationefficientlearningdeepnetworks}, FedProx \cite{li2020federatedoptimizationheterogeneousnetworks}, FedMA \cite{wang2020federatedlearningmatchedaveraging}, FedDyn \cite{acar2021federatedlearningbaseddynamic}, and SCAFFOLD \cite{karimireddy2021scaffoldstochasticcontrolledaveraging}, aim to develop generalized global models. Recently, personalized FL has gained attention, exploring methods based on multi-task learning \cite{10.5555/3294996.3295196}, interpolation and fine-tuning \cite{deng2020adaptivepersonalizedfederatedlearning}, meta-learning \cite{10.5555/3495724.3496024}, regularization \cite{li2021dittofairrobustfederated}, and knowledge distillation \cite{wu2024fedcacheknowledgecachedrivenfederated}. However, these methods often struggle to balance generalization and personalization, which is critical in real-world scenarios where both in-distribution and out-of-distribution classes can appear during inference.

\subsection{Personalized Federated Learning (PFL)}

PFL arises to address the challenges posed by statistical heterogeneity and non-IID data distributions in traditional FL. Conventional FL methods, such as FedAvg, often experience decreases in accuracy when training on non-IID data, mainly due to client drift. Tan et al. \cite{Tan_2023} categorize PFL approaches into two main strategies: 
\begin{inparaenum}
    \item personalization through a global model, which involves training a single global model, and 
    \item developing personalized models, where individual models are trained specifically for each client.
\end{inparaenum}

Most personalization techniques for global FL models \cite{sim2019investigationondevicepersonalizationendtoend} have a two-step process. The first step is training a global model together across clients, and the second step is refining the global model using each client’s private data to tailor it to their local data distribution \cite{9210355}. Typically, PFL methods employ FedAvg as the default training algorithm during global model training in FL settings, followed by additional local adaptation steps to personalize the model for individual clients. Alternatively, the personalized model learning strategy modifies the FL aggregation process and employs diverse learning algorithms to achieve PFL. These strategies can be further classified into architecture-based and similarity-based methods. The objective of this approach is to collaboratively train personalized models for a group of clients, leveraging their non-IID data distributions while simultaneously preserving their privacy \cite{Firdaus}. 

\subsection{Knowledge Distillation in Federated Learning}
Knowledge distillation (KD) \cite{hinton2015distillingknowledgeneuralnetwork} involves training a simpler student model to follow the predictions of a more complex teacher model. This approach ensures that the student model consistently aligns its outputs with the teacher's, effectively transferring knowledge. The difference between their predictions, often measured using KL divergence (soft loss), is the primary objective for training the student. In federated settings, distillation-based approaches \cite{li2019fedmdheterogenousfederatedlearning, lin2021ensembledistillationrobustmodel, jeong2023communicationefficientondevicemachinelearning, wu2024fedcacheknowledgecachedrivenfederated} use model distillation techniques, or knowledge transfer methods, to share insights by exchanging output logit vectors instead of model parameters. For instance, FedDistill \cite{jeong2023communicationefficientondevicemachinelearning} offers an alternative to centralizing local models. Aggregating soft scores rather than local models enables the creation of personalized models with diverse architectures. Similarly, FedCache \cite{wu2024fedcacheknowledgecachedrivenfederated} leverages hashes and labels to identify the R nearest neighbors within the same class using Hierarchical Navigable Small World (HNSW). It aggregates the collected soft labels based on the hash similarity of the samples. It sends the resulting aggregated logits back to the local nodes, allowing them to update their models with the latest global aggregation update. However, the assistance of labels on the server could potentially lead to privacy leakage.

\begin{figure*}[htbp]
    \centering
    \includegraphics[width=0.9\linewidth]{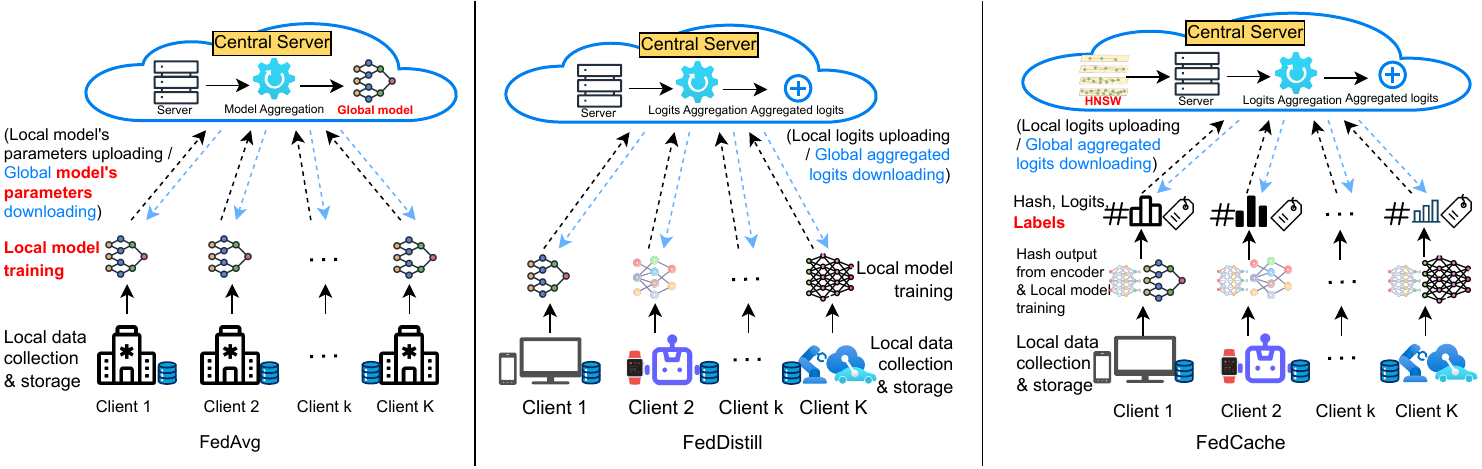}
    \caption{Overview of federated learning frameworks with edge devices, highlighting the key distinguishing features of FedAvg, FedDistill, and FedCache (marked in red). FedAvg necessitates homogeneous model architectures and faces challenges such as high communication overhead and privacy concerns due to the sharing of model parameters. FedDistill relies on coarse-grained logit exchange, which is insufficient for achieving effective personalized federated learning (PFL). FedCache involves sharing client-labels, raising privacy risks, and lacks a hierarchical structure to control the granularity of knowledge exchange.
    }
    \label{fig:framework}
\end{figure*}

\subsection{Federated Learning for Generalized Optimization}

Global optimization methods in FL aim to develop a model generalising data from all participating clients. These methods can be categorized into these approaches:

\subsection*{1) Approaches to Mitigate Client Drift}

To address client drift during the local training phase, various methods leverage global knowledge to guide local updates. Weight-based methods \cite{li2020federatedoptimizationheterogeneousnetworks, gao2022feddcfederatedlearningnoniid} use proximal terms to minimize discrepancies between local and global models or apply drift correction factors to address parameter variations. Feature-based methods \cite{li2021modelcontrastivefederatedlearning} aim to align the outputs of clients in latent spaces or confine client learning to similar representations by penalizing inconsistencies through feature contrast. Despite their potential, these methods often fail to address feature heterogeneity, limiting performance improvements. Prediction-based approaches \cite{lee2022preservationglobalknowledgenottrue, han2022fedxunsupervisedfederatedlearning} rely on public datasets and aggregate soft-label predictions instead of model parameters or gradients. This reduces communication overhead and facilitates knowledge distillation. 

\subsection*{2) Utilizing Auxiliary Data for Training}

Given the heterogeneity of client data, local models often struggle to generalize to missing classes. To address this, researchers have explored:
\begin{inparaenum}
    \item Sharing public datasets \cite{li2019fedmdheterogenousfederatedlearning},
    \item Generating synthetic data \cite{jeong2023communicationefficientondevicemachinelearning, hao2021fairfederatedlearningzeroshot}, or
    \item Using truncated versions of private datasets \cite{guha2019oneshotfederatedlearning}.
\end{inparaenum}

While these approaches improve performance, they raise privacy concerns as raw or synthetic data can expose sensitive information.

\subsection{Federated Learning for Personalized Optimization}

PFL aims to develop tailored models for individual clients based on their requirements and data characteristics. These methods \cite{Tan_2023, mansour2020approachespersonalizationapplicationsfederated} enhance the global model’s generalization capability before fine-tuning it with local client data. Data-based techniques use data augmentation to mitigate statistical heterogeneity \cite{hao2021fairfederatedlearningzeroshot, 9857319, 9141436} or adaptively sample subsets to accelerate convergence \cite{9155494, pmlr-v151-jee-cho22a}. However, sharing augmented data introduces privacy risks, and {Model-based techniques} regulate the personalization of local models by leveraging global knowledge \cite{Fallah, dinh2022personalizedfederatedlearningmoreau}. For example, \textbf{PLGU}\cite{10204784} integrates universal knowledge to optimize personalization while preventing clients from experiencing suboptimal performance.

\section{Methodology}
\subsection{Problem Formulation}
We consider an FL system with $N$ clients, denoted as 
$\mathcal{C} = \{C_i\}_{i=1}^N$, each having a local dataset $\mathcal{D}_i = \{X_i, Y_i\}$. The goal of the FL system is to collaboratively train a global model $\theta$ by aggregating the local updates $\theta_i$ 
from the clients. Each client $C_k$ also maintains a local model $M_k$. The central server $S$ coordinates the training process by collecting local updates and combining them to refine the global model.

In traditional FL\cite{mcmahan2023communicationefficientlearningdeepnetworks}, clients collectively aim to find a model parameter vector $\mathbf{w} \in \mathbb{R}^n$ that maps input $\mathbf{x}$ to output label $\mathbf{y}$, such that $\mathbf{w}$ minimizes the global loss. The global objective is expressed as:

\begin{equation}
F(\mathbf{w}) = \sum_{k=1}^N p_k F_k(\mathbf{w}),
\end{equation}

where $p_k$ is the proportion of data held by client $k$, and $F_k(\mathbf{w})$ is the local objective function for client $k$, defined as:

\begin{equation}
F_k(\mathbf{w}) = \frac{1}{|\mathcal{B}_k|} \sum_{\xi \in \mathcal{B}_k} f(\mathbf{w}, \xi),
\end{equation}

where $\mathcal{B}_k$ is the local dataset for client $k$, containing $|\mathcal{B}_k|$ data samples drawn from its local distribution $D_k$. The function $f(\mathbf{w}, \xi)$ is the composite loss for an individual data point $\xi$.

Due to high data heterogeneity across clients, the globally optimized parameter vector $\mathbf{w}^*$ that minimizes the global objective $F(\mathbf{w})$ may not generalize well for clients whose local objectives $F_k(\mathbf{w})$ deviate considerably from the global model. Such clients may opt to train their personalized models $\mathbf{w}_k \in \mathbb{R}^{n_k}$ by focusing solely on their local objectives. While this approach is effective for clients with abundant training data (where the local distribution $\mathcal{D}_k$ is well-represented by the empirical distribution $\hat{\mathcal{D}}_k$), it struggles for clients with limited data. For such clients, the mismatch between $\mathcal{D}_k$ and $\hat{\mathcal{D}}_k$ results in poor generalization and diminished benefits from FL participation. Additionally, standard FL assumes all clients use identical model architectures, which often fails to account for real-world device variability. This heterogeneity leads to a mismatch between model sizes and architectures across clients.

PFL addresses these challenges by adapting the global model $M_G$ to better align with the specific data distributions $\mathcal{D}_k$ of individual clients while leveraging the collective knowledge of all participants.

\begin{figure*}[htbp]
    \centering
    \includegraphics[width=0.8\linewidth]{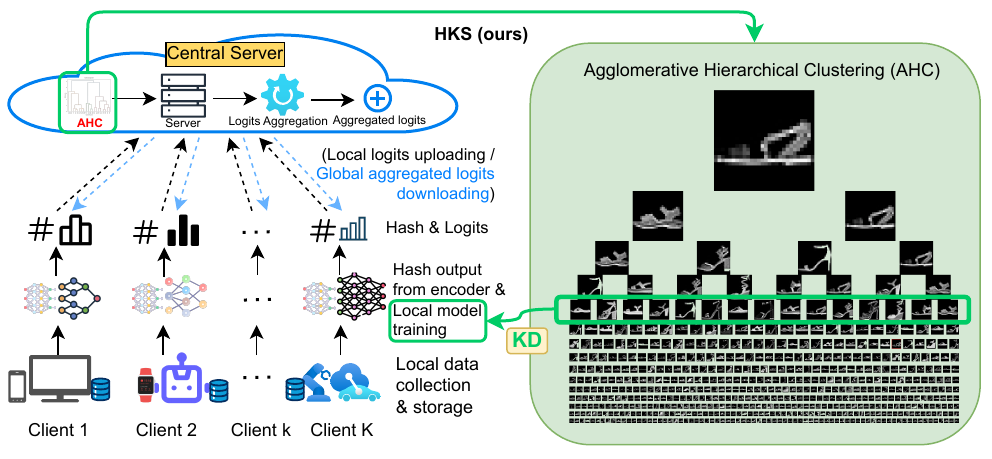}
    \caption{Overview of the HKS framework.}
    \label{fig:hks_framework}
\end{figure*}

\subsection{Knowledge Distillation}
Given a sample $\xi$, consider a student network $M_s$ and a teacher network $M_t$. Knowledge distillation involves transferring the teacher’s knowledge to the student by aligning their softened probability distributions. The softened probabilities are calculated as:
\begin{equation}
q = \text{softmax}\left(\frac{z}{T}\right),
\end{equation}
where $z$ is the logit output of $M(\xi)$, and $T$ is the temperature factor controlling the smoothness of $q$. The alignment loss helps the student network learn the relative confidence of the teacher network across categories, enhancing generalization \cite{10446765}. A common alignment loss is the Kullback-Leibler (KL) divergence:
\begin{equation}
L_{\text{Align}}(w_s, w_t; x) = L_{\text{KL}}(q_s, q_t) = -\sum_{i=1}^C q_t(i) \log\left[\frac{q_s(i)}{q_t(i)}\right].
\end{equation}

where $C$ is the number of categories.

\subsection{Hierarchical Knowledge Structuring (HKS)}
To address these challenges, we propose \textbf{Hierarchical Knowledge Structuring (HKS)}, which introduces a bottom-up hierarchical clustering mechanism on the server. The framework aims to:
\begin{enumerate}
    \item allow different levels of personalization to mitigate the effects of data heterogeneity; and
    \item accommodate heterogeneous devices with varying model architectures by sharing logits instead of model parameters.
\end{enumerate}
HKS leverages \textit{knowledge distillation} to guide local learning, addressing data imbalance and overfitting issues.

\subsection{HKS Training Framework}
As illustrated in Figure~\ref{fig:hks_framework}, the HKS training process consists of the following key steps:
\begin{enumerate}
    \item \textbf{Local Training:} Each client performs local training for 10 warm-up epochs.
    \item \textbf{Logits Upload:} After local training, each client uploads its logits to the server.
    \item \textbf{Clustering:} The server performs agglomerative hierarchical clustering on the received logits and aggregates them within each cluster.
    \item \textbf{Clustered Knowledge Sharing:} The aggregated logits are shared with the corresponding clients.
    \item \textbf{Knowledge Distillation:} Each client performs local training while distilling knowledge from the new logits.
\end{enumerate}

\subsection{Personalized Federated Learning with HKS}
\textbf{Logit Clustering:} The global server aggregates logits from all clients and applies a bottom-up clustering approach to structure them hierarchically. Each level of the hierarchy represents logits at varying levels of granularity:
\begin{itemize}
    \item \textbf{Per-sample logits:} Fine-grained knowledge tailored to individual samples.
    \item \textbf{Per-class logits:} Intermediate knowledge generalized across specific classes.
    \item \textbf{Global logits:} Coarse-grained knowledge capturing global patterns across all clients.
\end{itemize}

\textbf{Knowledge Distillation:} During local training, each client incorporates logits from the hierarchical structure:
\begin{itemize}
    \item Fine-grained logits for personalized insights.
    \item Coarse-grained logits to encourage global generalization.
\end{itemize}

The HKS framework balances generalization and personalization by enabling knowledge transfer at multiple levels of granularity. Each client's training objective is formulated as:
\begin{equation}
\mathcal{L}_{\text{total}} = \mathcal{L}_{\text{local}} + \alpha 
\cdot
\mathcal{L}_{\text{distill}},
\end{equation}

where $\mathcal{L}_{\text{local}}$ is the local training loss, $\mathcal{L}_{\text{distill}}$ is the knowledge distillation loss, and $\alpha$ is a weighting factor.

By integrating hierarchical clustering and knowledge distillation, HKS ensures that even clients with limited data can benefit from participating in FL, improving both generalization and personalization.

\begin{algorithm}[ht]
\caption{HKS: Hierarchical Knowledge Structuring}
\begin{algorithmic}[1]

\STATE \textbf{Input:} Client datasets $\{D_k\}_{k=1}^N$, number of classes $C$, temperature $T$, balancing factor $\alpha$, pre-trained encoder $E$, warmup epochs $W$

\STATE \textbf{Output:} Personalized models $\{M_k\}_{k=1}^N$

\STATE Initialize KnowledgeCache $\mathcal{K}$ with $C$ classes
\STATE Initialize client models $\{M_k\}_{k=1}^N$

\FOR{each client $k$ in parallel}
    \STATE Get feature hashes $\{h_i\}$ for samples using encoder $E$
    \STATE Add hashes to $\mathcal{K}$
\ENDFOR

\STATE Build hash relations in $\mathcal{K}$ using HNSW index

\FOR{each epoch $t$}
    \FOR{each client $k$ in parallel}
        \FOR{each batch $(x_i, y_i)$ in $D_k$}
            \STATE Forward pass: $\ell_i = M_k(x_i)$
            \STATE Compute cross-entropy loss $L_{CE}(\ell_i, y_i)$
            
            \IF{$t \geq W$}
                \STATE Update $\mathcal{K}$ with current logits $\ell_i$
                \STATE Fetch teacher knowledge based on granularity:
                \STATE \quad ``top": highest level cluster
                \STATE \quad ``middle": mid-level cluster
                \STATE \quad ``bottom": lowest level cluster
                \STATE \quad ``all": knowledge from all levels
                \STATE Compute KD loss $L_{KD}$ with temperature $T$
                \STATE Total loss: $L = L_{CE} + \alpha L_{KD}$
            \ELSE
                \STATE Total loss: $L = L_{CE}$
            \ENDIF
            
            \STATE Update model parameters using SGD
        \ENDFOR
    \ENDFOR
    
    \IF{$t \geq W$}
        \STATE Perform hierarchical clustering on $\mathcal{K}$ to $C$ clusters
        \STATE Update cluster assignments and paths
    \ENDIF
\ENDFOR

\end{algorithmic}
\end{algorithm}

\section{Experiments and Evaluations}
Our evaluation aims to address the following questions:
\begin{itemize}
    \item Can the proposed \textbf{HKS} framework achieve superior personalized or generalized performance compared to baseline methods like \textbf{FedDistill} and \textbf{FedCache}?
    \item How does data heterogeneity, controlled by partitioning coefficients ($\alpha$), impact personalized and generalized model performance?
\end{itemize}

\subsection{Experimental Setup}

We evaluate our proposed \textbf{HKS} framework using the \textbf{FashionMNIST} dataset, a widely-used benchmark for image classification with 10 classes. The dataset comprises 60,000 training images and 10,000 test images of clothing items distributed across 20 clients in an FL environment. To simulate non-IID data distributions, we partition the dataset using a Dirichlet distribution with partitioning coefficients ($\alpha$) of 1.0 and 0.5, representing moderate and high levels of data heterogeneity, respectively.

The FL setup runs 18 communication rounds, with each client performing 1 local training epoch per round using SGD as the optimizer (learning rate = 0.01, batch size = 8). The distillation loss weight ($\alpha$) in the knowledge distillation loss formulation is set to 1.5 by default, balancing local and distillation-based learning.

\subsubsection{Model Heterogeneity}

To simulate realistic deployment scenarios with diverse device capabilities, we implement three different ResNet architectures of varying capacities:

\begin{itemize}
    \item \textbf{ResNet-8}: A lightweight model with [1,1,1] basic blocks, suitable for resource-constrained devices.
    \item \textbf{ResNet-16}: A medium-capacity model with [2,2,2] basic blocks, for moderate-capability devices.
    \item \textbf{ResNet-20}: A full-capacity model with [3,3,3] basic blocks, for high-capability devices.
\end{itemize}

Client models are assigned following a systematic pattern where $i$ is the index of the client:
\begin{itemize}
    \item Every client with $i$ modulus 3 = 0 receives a ResNet-8.
    \item Every client  with $i$ modulus 3 = 1 receives a ResNet-16.
    \item Every client with $i$ modulus 3 = 2 receives a ResNet-20.
\end{itemize}

This distribution ensures a balanced mix of device capabilities across the federated network while reflecting real-world heterogeneity in computational resources.

\begin{figure}[!t]
    \centering
    \includegraphics[width=0.9\linewidth]{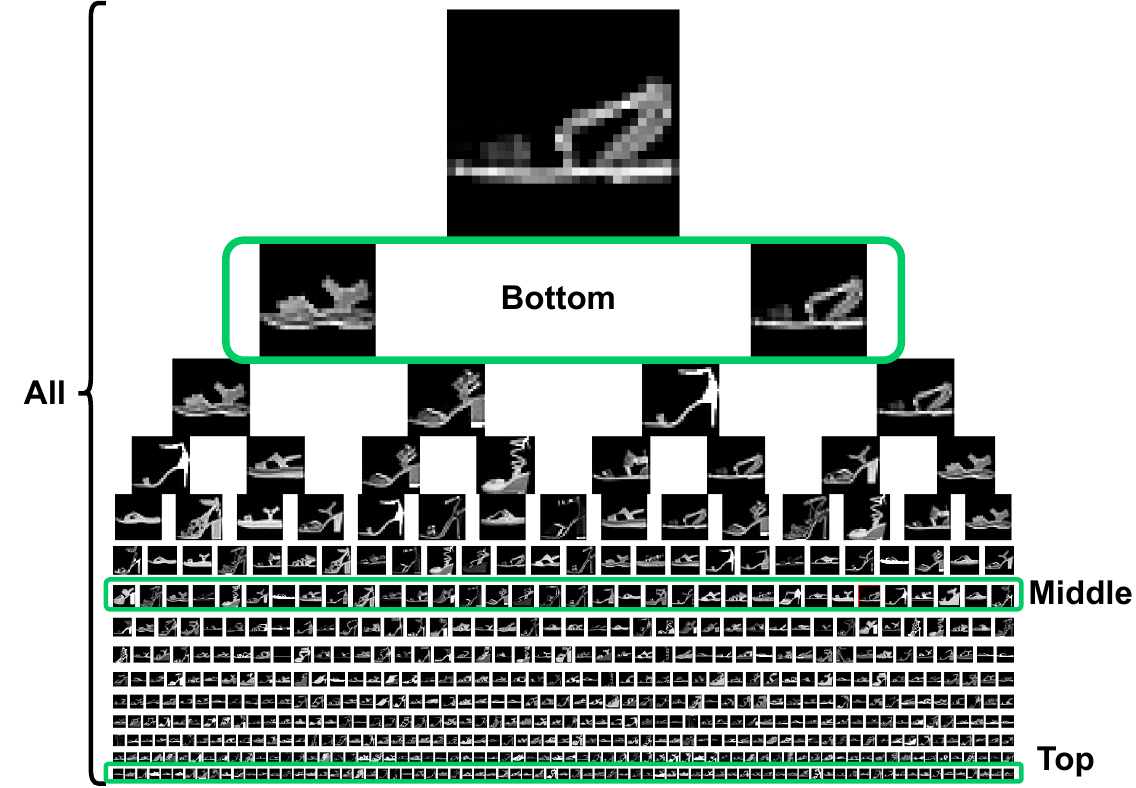}
    \caption{Cluster path of a target sample in the bottom-up hierarchical clustering. The top-most node represents the initial cluster containing the target sample. Each subsequent row represents a merged cluster showing member samples.}
    \label{fig:cluster_path}
\end{figure}

\subsubsection{Data Distribution}

To simulate non-IID data distributions, we partition the dataset using a Dirichlet distribution with partitioning coefficients ($\alpha$) of 1.0 and 0.5, representing moderate and high levels of data heterogeneity, respectively. This setup allows us to evaluate our method's effectiveness under varying degrees of data distribution skew across heterogeneous devices.

\begin{table*}[htbp]
\centering
\caption{Results for Personalized and Generalized Metrics with $\alpha = 1.0$}
\label{tab:results_alpha_1}
\begin{tabular}{@{}ccccc@{}}
\toprule
\textbf{Method} & \textbf{Hyperparameters} & \textbf{Personalized (MAUA)} & \textbf{Generalized (Global Test Accuracy)} \\ \midrule
\multirow{3}{*}{FedAvg} & ResNet-8 & 49.53 & 44.06 \\
& ResNet-16 & 42.24 & 36.79 \\
& ResNet-20 & 44.71 & 31.35 \\ \cmidrule{1-4} 
FedDistill & - & 86.67 & 67.22 \\ \cmidrule{1-4} 
\multirow{5}{*}{FedCache} 
  & $R = 1$ & 87.49 & 68.17 \\ 
&  $R = 4$ & 87.44 & 68.88 \\ 
&  $R = 16$ & 87.55 & 68.26 \\ 
&  $R = 64$ & 87.87 & \textbf{69.57} \\ 
&  $R = 256$ & 87.52 & 68.87 \\ \cmidrule{1-4} 
\multirow{4}{*}{\textbf{HKS (ours)}} 
  & Granularity: \textit{top} & 86.96 & 68.59 \\
&  Granularity: \textit{middle} & \underline{89.16} & 68.30 \\
&  Granularity: \textit{bottom} & 88.44 & 68.31 \\
&  Granularity: \textit{all} & \textbf{89.38} & \underline{69.00} \\ 
\bottomrule
\end{tabular}
\begin{tablenotes}
\small
\item[*] \centering For $\alpha = 1.0$, the highest values are shown in \textbf{bold} and the second-highest values are \underline{underlined}.
\end{tablenotes}
\end{table*}

\begin{table*}[htbp]
\centering
\caption{Results for Personalized and Generalized Metrics with $\alpha = 0.5$}
\label{tab:results_alpha_0.5}
\begin{tabular}{@{}ccccc@{}}
\toprule
\textbf{Method} & \textbf{Hyperparameters} & \textbf{Personalized (MAUA)} & \textbf{Generalized (Global Test Accuracy)} \\ \midrule
\multirow{3}{*}{FedAvg} & ResNet-8 & 20.91 & 13.88 \\
& ResNet-16 & 22.44 & 20.89 \\
& ResNet-20 & 24.81 & 23.19 \\ \cmidrule{1-4} 
FedDistill & - & 89.76 & 58.72 \\ \cmidrule{1-4} 
\multirow{5}{*}{FedCache} 
  & $R = 1$ & 89.20 & \underline{59.49} \\
&  $R = 4$ & 89.45 & 59.03 \\
&  $R = 16$ & 89.57 & 59.15 \\
&  $R = 64$ & 89.68 & 58.73 \\
&  $R = 256$ & 89.38 & 58.74 \\ \cmidrule{1-4} 
\multirow{4}{*}{\textbf{HKS (ours)}} 
  & Granularity: \textit{top} & 89.43 & \textbf{59.75} \\
&  Granularity: \textit{middle} & \textbf{90.79} & 56.09 \\
&  Granularity: \textit{bottom} & \underline{90.61} & 58.26 \\
&  Granularity: \textit{all} & 90.56 & 56.89 \\ 
\bottomrule
\end{tabular}
\begin{tablenotes}
\small
\item[*] \centering For $\alpha = 0.5$, the highest values are shown in \textbf{bold} and the second-highest values are \underline{underlined}.
\end{tablenotes}
\end{table*}

\subsubsection{Hyperparameter}

For \textbf{FedCache}, results for various related sample settings ($R \in \{1, 4, 16, 64, 256\}$) are included. The hyperparameter $R$ controls the number of nearest neighbors of the same class from which the logits are aggregated for knowledge distillation. A larger $R$ allows more information from a wider range of neighboring samples to be distilled.

For \textbf{HKS}, results are reported for different granularity levels (\textit{top}, \textit{middle}, \textit{bottom}, \textit{all}) illustrated in Figure~\ref{fig:cluster_path}. The granularity hyperparameter defines how logits are aggregated from the hierarchical structure:
\begin{itemize}
    \item \textit{Top}: The logits are aggregated at the highest level of the hierarchy, where the number of clusters equals the number of classes. This prioritizes sharing generalized knowledge across all clients.
    \item \textit{Middle}: Aggregation occurs at the middle layer of the hierarchy, balancing between general and specific knowledge.
    \item \textit{Bottom}: Aggregation is performed at the first level of the hierarchy, emphasizing personalized knowledge by leveraging finer-grained, client-specific clusters.
    \item \textit{All}: The KD loss is averaged across all clusters in the cluster path of a sample, combining information from all hierarchical levels.
\end{itemize}

\subsection{Baselines and Metrics}

We compare HKS against three baseline methods:
\begin{itemize}
    \item \textbf{FedAvg}\cite{mcmahan2023communicationefficientlearningdeepnetworks}: The standard FL algorithm that averages model parameters from clients to update the global model.
    \item \textbf{FedDistill}\cite{jeong2023communicationefficientondevicemachinelearning}: A distillation-based framework that transfers client knowledge via sample-grained logits with a public dataset or class-grained logits.
    \item \textbf{FedCache}\cite{wu2024fedcacheknowledgecachedrivenfederated}: An enhanced distillation method that leverages cached logits to reduce communication cost.
\end{itemize}

The evaluation metrics include:
\begin{itemize}
    \item \textbf{Personalized Model Accuracy (MAUA)}\cite{Mills2020MultiTaskFL}: Maximum average user model accuracy across communication rounds on clients' local datasets, measuring the effectiveness of personalization across different device capabilities.
    \item \textbf{Generalized Model Accuracy}: Average global test accuracy across all clients on an evenly distributed global dataset, indicating methods' ability to maintain uniform performance across heterogeneous devices.
\end{itemize}

\subsection{Experimental Results and Analysis}
Table \ref{tab:results_alpha_1} presents the results of the experiments conducted on the \textbf{FashionMNIST} dataset \cite{xiao2017fashionmnistnovelimagedataset}, highlighting the performance of the proposed \textbf{HKS} framework compared to baseline methods (\textbf{FedAvg} \cite{mcmahan2023communicationefficientlearningdeepnetworks}, \textbf{FedDistill} \cite{jeong2023communicationefficientondevicemachinelearning} and \textbf{FedCache} \cite{wu2024fedcacheknowledgecachedrivenfederated}). The results present both personalized accuracy (MAUA) \cite{Mills2020MultiTaskFL} and generalized accuracy (i.e., global test accuracy) for the compared methods. Results are reported under two settings of data heterogeneity controlled by the partition coefficient $\alpha$ (1.0 and 0.5), after 18 communication rounds. Table \ref{tab:results_alpha_0.5} presents the results for the same metrics under the more challenging data heterogeneity setting with $\alpha = 0.5$. The performance results from the compared methods under varying levels of data heterogeneity are as follows:

\begin{enumerate}[leftmargin=*]
    \item  \textbf{FedAvg}: struggles notably in environments with high data diversity, such as when $\alpha = 0.5$. Both personalized and global accuracy drop considerably because model parameters are averaged across clients with imbalanced data to align the divergent updates. This naive averaging causes the global models to struggle with adapting to the distinct characteristics of each client’s local dataset. Its low MAUA and global test accuracy further emphasize its inability to adapt effectively to local data variations. In $\alpha = 1.0$ setting, deeper models like ResNet-20 overfit to local data distribution as they capture intricate patterns, leading to poorer generalization (31.35\% global accuracy) because deeper models’ overfitted parameters misalign during aggregation, which exacerbates model divergence. In contrast, shallow models (ResNet-8/16) alleviate overfitting by learning simpler features, performing better on global tests (44.06\% / 36.70\%). All models struggle due to extreme non-IIDness at $\alpha = 0.5$ because skewed data amplifies the conflicting parameter updates, which results in a global model that poorly aligns with any client's data distribution. As local models are initialized from a global model that is updated via parameter averaging, if the global model is poor due to aggregation issues (i.e., parameter divergence), local training starts from a suboptimal base, limiting its ability to learn effectively on local data (i.e., low MAUA). 
    \item \textbf{FedDistill}: performs better in more balanced settings, such as when $\alpha = 1.0$, where it achieves a high MAUA of 86.67\% and a global test accuracy of 67.22\%. Transferring knowledge through logits helps mitigate the effects of data heterogeneity. However, its global accuracy declines to 58.72\% in the $\alpha = 0.5$ setting, although it achieves a strong MAUA of 89.76\%. This performance decay indicates that while FedDistill can effectively leverage local knowledge, it may struggle to generalize in high data heterogeneity.
    \item \textbf{FedCache}: benefits from aggregating logits from multiple neighbors within the same class where more logit aggregation with higher hyperparameter value \( R \) generally improve global test accuracy. For example, with \( R = 64 \), FedCache achieves the highest global test accuracy of 69.57\% when \( \alpha = 1.0\). The MAUA values for FedCache remain good across different \( R \) values, indicating that it effectively balances personalized and generalized accuracy. However, larger \( R \) (e.g., 256) slightly reduces performance 68.87\% as this introduces noisy or less relevant neighbors from the HNSW graph, diluting the distilled knowledge's quality. Moreover, in the more challenging setting of \( \alpha = 0.5\), the global test accuracy is lower due to intra-class noise, introduced by blending dominant local features with sparse/biased external features, diluting knowledge quality (e.g., 90\% ``sandals" in one client while 5\% in another), as the combined logits blend conflicting information.  
    \item \textbf{HKS}: consistently achieves higher personalized MAUA across both partition settings ($\alpha=1.0$ and $\alpha=0.5$), reflecting its strength in tailoring models to client-specific data distributions. For global test accuracy, HKS outperforms baseline methods in the more challenging heterogeneity setting ($\alpha=0.5$) and achieves comparable results to FedCache in an easier scenario (i.e., $\alpha=1.0$). In particular, the ``all" granularity achieves the highest accuracy compared to other granularity at ($\alpha=1.0$), combining broad and fine-grained features to maximize generalization and personalization. However, clients develop unique feature specialisations at ($\alpha=0.5$). Aggregating all levels mixes conflicting cluster knowledge, slightly lowering performance. Hence, targeted granularity (e.g., ``middle") outperforms ``all" in non-IID settings by avoiding conflicting clusters. Since HKS clustering is class-agnostic, it could group similar features between classes, which is more robust to class imbalance.  
\end{enumerate}

When the data is highly heterogeneous (non-IID, e.g., $\alpha = 0.5$), each client primarily trains on a narrow subset of the overall data distribution (e.g., specific classes or features). This setting allows local models to ``over-specialize" to their private data. The model performs exceptionally well on its local testing data because the training and testing datasets are from the same subset, leading to higher MAUA. However, the lack of exposure to a diverse data distribution causes poorer performance on the global test dataset, where classes are more evenly distributed.

In contrast, less heterogeneous settings ($\alpha=1.0$) enable more effective global knowledge sharing, as clients' data distributions align more closely with the global distribution, i.e., class samples are more evenly spread across clients instead of being skewed on a few clients. The broader class overlap enables more meaningful knowledge transfer between clients. KD acts as a regularizer, enhancing shared understanding of the global data distribution and improving generalization.

Analysis of different granularity levels in \textbf{HKS} reveals trade-offs between personalization and generalization. At the \textit{top} level, knowledge sharing occurs across all clients, promoting better generalization as evidenced by higher global test accuracy (\textbf{68.59\%} for $\alpha=1.0$ and \textbf{59.75\%} for $\alpha=0.5$) but potentially sacrificing some personalization benefits. The \textit{middle} granularity level achieves the strongest personalization performance (MAUA of \textbf{89.16\%} for $\alpha=1.0$ and \textbf{90.79\%} for $\alpha=0.5$) by balancing local specialization with moderate knowledge sharing among similar clients. The \textit{bottom} granularity restricts knowledge sharing to very similar clients, maintaining strong personalization (\textbf{88.44\%} and \textbf{90.61\%} respectively) while still outperforming baseline methods.

The choice of optimal granularity level depends on the data heterogeneity. In less heterogeneous settings, using all granularity levels (\textit{all}) achieves the best overall performance (\textbf{89.38\%} MAUA and \textbf{69.00\%} global accuracy), suggesting that multiple levels of knowledge sharing are beneficial when client data distributions are similar. However, in highly heterogeneous settings, different granularity levels show more pronounced trade-offs: the \textit{top} level excels in generalization (\textbf{59.75\%} global accuracy) while the \textit{middle} level achieves better personalization (\textbf{90.79\%} MAUA).

These findings suggest that practitioners should select granularity levels based on their requirements and data characteristics. When generalization is essential or when working with highly heterogeneous data, \textit{top}-level granularity is recommended because it promotes broad knowledge sharing across all clients, mitigating over-specialization and improving global test accuracy. For applications prioritizing personalization, \textit{middle} granularity offers the best performance as it enables targeted knowledge sharing among similar clients, preserving local specialization while benefiting from shared insights. When data distributions are relatively uniform, enabling all granularity levels can provide the best of both worlds. This multi-level knowledge sharing takes full advantage of the uniform overlap across clients, maximizing personalization and generalization. These findings show that \textbf{HKS} can balance personalization and generalization trade-offs by leveraging knowledge at different granularity. Its hierarchical knowledge structuring achieves superior personalized performance and competitive global performance, especially under higher levels of heterogeneity, showing its effectiveness for FL in diverse settings.

\section{Conclusion and Future Work}
This study introduces the HKS framework as a new way to balance personalization and generalization in FL when faced with diverse devices and data. By structuring logits into a multi-level hierarchy, HKS facilitates more efficient knowledge sharing among clients with varying models and data. The framework is tailored to account for differences in data distributions across clients. An important advantage of the HKS approach is that it does not require access to client label information or a public dataset to achieve competitive performance. Instead, HKS utilizes hierarchical clustering to organize the knowledge exchange to improve learning. Additionally, this approach maintains communication costs at a low level, as only logits are exchanged between clients rather than full model parameters or data. Experimental results show that HKS consistently achieves superior personalized accuracy (MAUA), particularly in more heterogeneous settings, while demonstrating competitive global performance compared to state-of-the-art methods. HKS is a promising approach for FL systems in heterogeneous environments where privacy and communication costs are critical. Future work should focus on dynamic hyperparameter tuning to maintain robustness across varying non-IID scenarios.

\bibliographystyle{myIEEEtran} %
%\bibliography{references} %

\end{document}